\documentclass[article]{elsarticle}

\usepackage{hyperref}

\journal{Journal of \LaTeX\ Templates}
\usepackage{multirow}
\usepackage{booktabs}
\usepackage{amssymb}
\usepackage{amsthm}

\begin{document}

\begin{frontmatter}

\title{Talking Head Generation Driven by Speech-Related Facial Action Units and Audio- Based on Multimodal Representation Fusion}

\author{Sen Chen}
\author{Zhilei Liu\corref{mycorrespondingauthor}}
\ead{zhileiliu@tju.edu.cn}
\cortext[mycorrespondingauthor]{Corresponding author}
\author{Jiaxing Liu}
\author{Longbiao Wang}
\address{College of Intelligence and Computing, Tianjin University, Tianjin, China.}

\begin{abstract}
Talking head generation is to synthesize a lip-synchronized talking head video by inputting an arbitrary face image and corresponding audio clips. Existing methods ignore not only the interaction and relationship of cross-modal information, but also the local driving information of the mouth muscles. In this study, we propose a novel generative framework that contains a dilated non-causal temporal convolutional self-attention network as  a multimodal fusion module to promote the relationship learning of cross-modal features. In addition, our proposed method uses both audio- and speech-related facial action units (AUs) as driving information. Speech-related AU information can guide mouth movements more accurately. Because speech is highly correlated with speech-related AUs, we propose an audio-to-AU module to predict speech-related AU information. We utilize pre-trained AU classifier to ensure that the generated images contain correct AU information. We verify the effectiveness of the proposed model on the GRID and TCD-TIMIT datasets. An ablation study is also conducted to verify the contribution of each component. The results of quantitative and qualitative experiments demonstrate that our method outperforms existing methods in terms of both image quality and lip-sync accuracy.
\end{abstract}



\end{frontmatter}


\section{Introduction}\label{sec:introduction}
Face animation synthesis has attracted increasing attention in academic and industrial fields, and is considered essential in the real-life applications of human-computer interaction, online teaching, film making, virtual reality, and computer games, among others~\cite{liu2020synthesizing,chen2018lip,sunspeech2talking}. Traditionally, facial synthesis in computer-generated imagery (CGI) has been performed using face capture methods. Although these methods have been rapidly developed in the past few years, they often require massive resource expenditure. Automatic face synthesis remains a challenge in the field of computer graphics. Recently, researchers have attempted to apply artificial intelligence technology to talking head generation. This research studies methods of generating talking head videos by inputting an arbitrary face image as an identity image and driving information related to mouth movement, for example, speech audio and text\cite{chen2020comprises}. 

Before deep learning became popular, many researchers relied on hidden Markov models (HMMs) to capture the dynamic relationship between audio and lip motion \cite{yamamoto1998lip,choi1999baum,xie2007coupled}. In recent years, deep neural networks (DNNs) have been widely used in talking head generation. Some methods selected the best-matched lip region image from a database of images of a specific person by inputting audio information and then synthesizing it into the target face \cite{journals/tog/SuwajanakornSK17,fan2015photo}. These methods are subject dependent and incur significant overhead when transferring to a new subject. Later, many researchers tried to generate arbitrary speakers instead of specific speakers \cite{DBLP:conf/aaai/Zhou000W19,chen2019hierarchical,chung2017you,zhou2020makelttalk}. However, these studies ignore the temporal relationship of the features when generating the image sequence, making the generated video incoherent. Some researchers have used a sequence modeling method to learn the temporal relationship of input features to solve this problem \cite{yu2020multimodal,song2018talking,vougioukas2019realistic,eskimez2020end}. Although these methods showed excellent results, the recurrent neural network (RNN) -based models had complicated structures, slow training speed, and could not be fully trained when the data were insufficient. Moreover, the essence of talking head generation is to convert the content information in the audio into accurate movements of the lips, which involves the information of the two modalities of vision and speech. All of the existing methods directly concatenate the features of the two modalities, which ignore the interaction and relationship of information between the different modalities. As a result, the information of the two modalities cannot be effectively used in prior models. 

Recent research has shown that temporal convolutional networks (TCN) \cite{bai2018empirical} can outperform recurrent networks in the task of dealing with sequential problems. The dilated convolution of TCN has a larger receptive field and can capture more temporal information in processing sequence problems. Because speech needs to make mouth preparations according to what will be said, the information in the next few frames has a positive effect on current mouth movement. Compared with the causal TCN proposed by Bai et al.\cite{bai2018empirical}, which does not use future information, the modified non-causal TCN \cite{farha2019ms} can utilize both past and future information. In attition, to promote relationship learning and fusion of cross-modal information, we need to integrate a self-attention mechanism into the non-causal TCN. Therefore, we propose a multimodal representation fusion module called dilated non-causal temporal convolutional self-attention network (TCSAN), which can simultaneously capture longer sequence information and learn the interrelationship between multimodal information.

In addition, talking is a kind of movement driven by facial muscles, especially in the mouth region. Regardless of which approach is adopted, the existing methods ignore the local driving information of the mouth muscles, resulting in a decrease in the synchronization between lip movement and audio. Inspired by this limitation, we use both audio and local information relating to the mouth muscles to drive talking head generation. 

Facial action coding system (FACS) is a comprehensive and objective description system of facial movements~\cite{Ekman}. It defines a set of basic facial action units (AUs), wherein each AU represents a basic facial muscle movement. FACS has attracted considerable attention in face editing \cite{pumarola2018ganimation,li2019face,liu2020region}, such as facial expression editing \cite{pumarola2018ganimation}. These studies have proven that AU information can be used to edit local facial regions. AU labels are usually extracted by facial image detection \cite{baltrusaitis2018openface,shao2018deep}; however, they can also be extracted through other modalities related to facial motion, such as speech audio. A few researchers have constructed the relationship between speech information and speech-related AU information in recent years \cite{meng2017listen,ringeval2015face}. They proved that the use of audio as information for speech-related AU recognition was feasible. Therefore, we propose an audio-to-AU module to obtain speech-related AU representation from speech as local driving information.

\begin{figure}[t]
  \centering
  \includegraphics[width=0.85\linewidth]{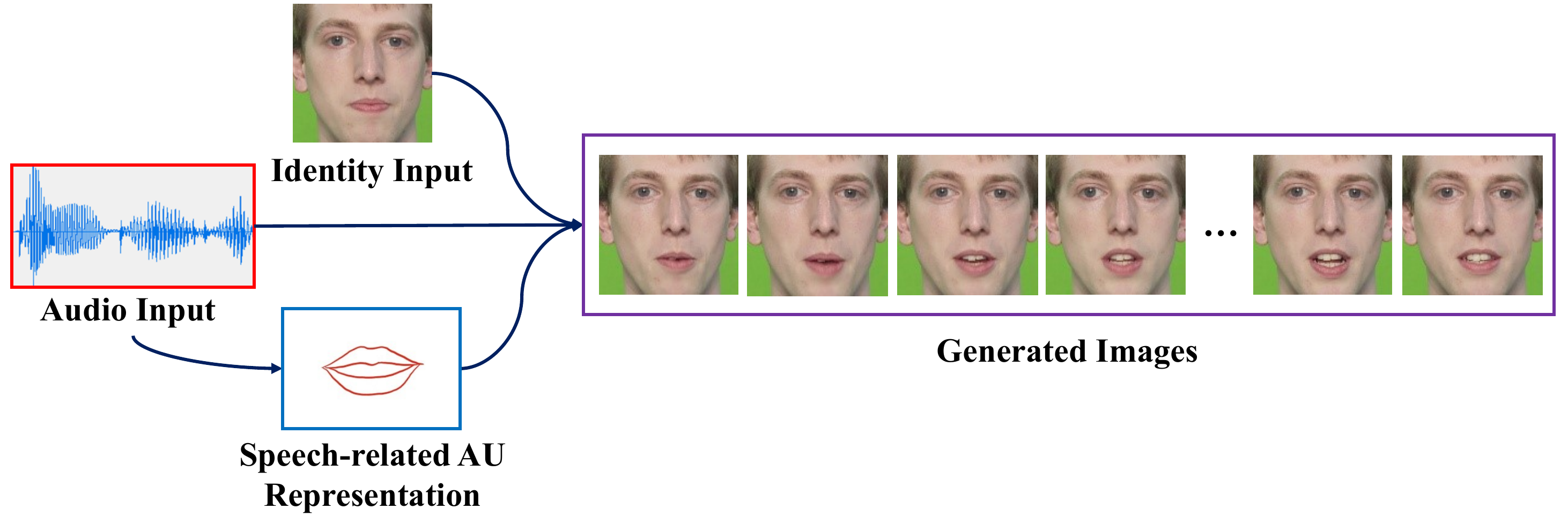}
  \caption {Proposed talking head generation system, which uses audio and speech-related AU information to drive a face image to generate speech video.}
  \label{fig.1}
\end{figure}

In summary, there are two main limitations in existing talking head generation methods. 1) The cross-modal features are simply concatenated and are not effectively used, and 2) the local information of the mouth muscles is ignored. To address these limitations, in this study, we propose a novel talking head generation method, which not only uses a non-causal temporal convolutional self-attention network to fuse multimodal features, but also uses speech-related facial action units as local information to drive the mouth muscle movements. As shown in Fig.~\ref{fig.1}, the proposed approach uses both audio and AU representations as the driving information to guide lip movement; hence, audio information drives the entire mouth, and AU information focuses on the local muscles. The contributions of our work can be summarized as follows.

\begin{itemize}

\item A novel talking head generation system is proposed by effectively fusing multimodal features, and integrating both audio and speech-related AUs as driving information to drive accurate mouth movement. 


\item A dilated non-causal temporal convolutional self-attention network (TCSAN) is proposed for multimodal fusion module to promote the interaction and relationship learning of cross-modal features; it can also learn the temporal relationship between consecutive frames well.

\item An Audio2AU module is designed in our framework to predict speech-related AU information from speech. Moreover, we add pre-trained AU classifier to supervise the model and ensure that the generated images contain correct AU information.

\item Extensive quantitative and qualitative experiments conducted on two datasets demonstrate that our framework achieves high-quality talking head generation of arbitrary identities and achieves significant improvement over existing methods.\sloppy

\end{itemize}

In comparison to the earlier conference version \cite{chen2021talking} of this work, we introduce the new temporal convolutional self-attention network in Sect. \ref{3.1} to promote cross-modal representation fusion effectively. Specifically, we introduce the working mechanism of our TCSAN module, and prove its advantages over the existing fusion methods in cross-modal representation learning and maintaining temporal relationship. With this improvement, the results generated by our talking head generation framework have better image quality and lip-sync accuracy compared to our earlier conference version\cite{chen2021talking}.

The remainder of this study is organized as follows. In Section 2, we describe relevant studies related to our work. In Section 3, we introduce our proposed framework in detail. In Section 4, we describe the datasets used in this study and the training details. In Section 5, extensive experiments are conducted for quantitative and qualitative evaluations. The final section concludes the work.

\section{Related Work}
\subsection{Audio-driven talking head generation}
This task uses audio as speech-related information to drive a facial image to generate a lip-synchronized talking face video. Before DNN was widely used, most methods for modeling the relationship between speech signal and lip motion were based on HMM \cite{bregler1997video,yamamoto1998lip,xie2007coupled}. Yamamoto et al. \cite{yamamoto1998lip} proposed a Succeeding-Viseme-HMM-based method to drive a lip movement sequence from an input acoustic speech signal. Xie et al. \cite{xie2007coupled} proposed a method based on a coupled hidden Markov model (CHMM) to explicitly model the subtle features of audio--visual speech, such as synchronization, asynchrony, and temporal coupling. Later, Suwajanakorn et al. \cite{journals/tog/SuwajanakornSK17} designed a DNN-based method to synthesize a talking head. They used a database of speeches by the then US President Obama to retrieve the lip region image that best matches the input audio and synthesized  it into a target image of Obama. Chung et al. \cite{chung2017you} realized the talking head generation of arbitrary identities based on an encoder--decoder structure. They used 350 ms audio frames as driving information, and each frame generated an image. Prajwal et al. \cite{prajwal2020lip} used a pre-trained SyncNet \cite{chung2016out} as a lip-sync discriminator for adversarial training with a generator. Chen et al.\cite{chen2020talking} and Zhou et al.\cite{zhou2021pose} proposed different methods to synthesize talking faces with controllable head poses to achieve natural head movements. Some studies have also attempted to generate specific emotional video portraits\cite{eskimez2021speech,ji2021audio}. Eskimez et al. \cite{eskimez2021speech} introduced an emotion encoder to generate a talking face video with a specific emotion, whereas our method generates talking face videos with specific AUs. Some researchers have used the sequence modeling method to enhance the temporal dependence of generated frames. Song et al. \cite{song2018talking} concatenated audio and identity features into a RNN to enhance the temporal dependence between frames. Vougioukas et al. \cite{DBLP:conf/bmvc/VougioukasPP18} proposed a temporal generative adversarial network (GAN) \cite{goodfellow2014generative}, which uses a gated recurrent unit (GRU) in the audio encoder and added Gaussian noise to generate spontaneous facial movements. However, 
these methods do not consider the interaction of cross-modal features and ignore the local information of the mouth muscles.

\subsection{Representation fusion}
Artificial intelligence is increasingly being applied to cross-domain problems. Many cross-domain studies have shown improvements in performance by fusing multimodal information \cite{pang2015deep,yan2016sparse,takahashi2017aenet,elmadany2018multimodal}. Therefore, it is of great significance to study the problems of complex cross-modal learning and modeling \cite{zhang2020multimodal}. Among the many fusion methods, attention-based fusion has become popular. 
Vaswani et al. \cite{vaswani2017attention} proposed the first attention-based sequence transformation model called transformer, which replaces the most commonly used recurrent layers in encoder--decoder architectures with multi-head self-attention. It was able to perform a large number of parallel calculations and learn long-distance dependencies better. The multi-head self-attention mechanism was quickly applied to other fields and achieved excellent results \cite{liu2021multimodal,zhang2019spatio}. Later, Li et al. \cite{li2020multimodal} proposed a multimodal fusion method based on a multi-head co-attention mechanism to improve the applicability of cross-modal data. Zhang et al. \cite{zhang2018image} proposed a channel attention mechanism to adaptively rescale channel-wise features by considering the interdependencies among channels when solving the image super-resolution problem. Wang et al. \cite{DBLP:conf/cvpr/WangWZLZH20} proposed an efficient channel attention (ECA) module able to generate channel attention through fast 1D convolution; the kernel size was adaptively determined by a nonlinear mapping of the channel dimension.

Existing talking head generation methods simply concatenate the cross-modal feature vectors, and then either directly generate the images through an image decoder \cite{prajwal2020lip}, or implicitly learn the relationship of cross-modal information through RNNs or Long Short-Term Memory networks (LSTMs) \cite{song2018talking,eskimez2020end}. These are not conducive to the relationship learning of cross-modal representations. The combination of self-attention mechanism and temporal convolutional networks \cite{bai2018empirical} can not only promote cross-modal representation fusion effectively, but also learn longer sequence information \cite{liu2020temporal}. Therefore, we combine multi-head self-attention with a non-causal TCN to fuse cross-modal representation.

\subsection{AU-based face editing}
As one of the most comprehensive ways to describe facial movements, the Facial Action Coding System (FACS) developed by Ekman and Friesen \cite{Ekman} has recently attracted widespread attention. Facial action units (AUs) defined by FACS can represent the contractions of specific facial muscles, and are widely used to edit the subtle movements of the face \cite{martinez2017automatic,pumarola2018ganimation,liu2020region}. Pumarola et al. \cite{pumarola2018ganimation} proposed an AU-based face-editing system that uses AU intensity labels to edit arbitrary face images and generate a face with specific facial muscle action. Liu et al. \cite{liu2020region} utilized local AU regional rules to control the status of each AU and used an attention mechanism to integrate them into the entire facial expression. Zhou et al. \cite{zhou2017photorealistic} developed a method to generate a face image with the target AU label using a conditional difference adversarial autoencoder. Facial action units are also used in face inpainting. For example, Li et al. \cite{li2019face} combined GAN and prior knowledge based on the dynamic structural information of facial action units to complete face images. By contrast, the proposed method utilizes speech-related AU information to edit the local region of the mouth.

\subsection{Speech-related AU recognition}
In addition to using face detection methods to perform AU recognition, as in most existing studies \cite{baltrusaitis2018openface,shao2018deep,wang2019weakly}, AU labels can also be extracted through other modalities related to facial motion. Because speech is highly correlated with speech-related AUs, a few researchers have attempted to construct models to establish the relationship between audio and speech-related AU information in recent years \cite{ringeval2015face,meng2017listen,meng2018improving}. As the first attempt to use acoustic cues for automatic AU detection, Ringeval et al. \cite{ringeval2015face} used low-level descriptors (LLD) of acoustic feature datasets to perform AU recognition through an RNN-LSTM model. Meng et al. \cite{meng2017listen} proposed a continuous-time Bayesian network (CTBN) to simulate the dynamic relationship between phonemes and AUs. Subsequently, AU recognition was performed by probabilistic inference using the CTBN model. Later, they also presented an audio--visual fusion framework based on a dynamic Bayesian network (DBN) \cite{meng2018improving}, which used both visual and audio channel information to recognize speech-related AUs. In this work, we propose an audio-to-AU module to predict speech-related AU information from all frames of a sequence of audio frames. 

\begin{figure*}[!t]
  \centering
  \includegraphics[width=0.9\linewidth]{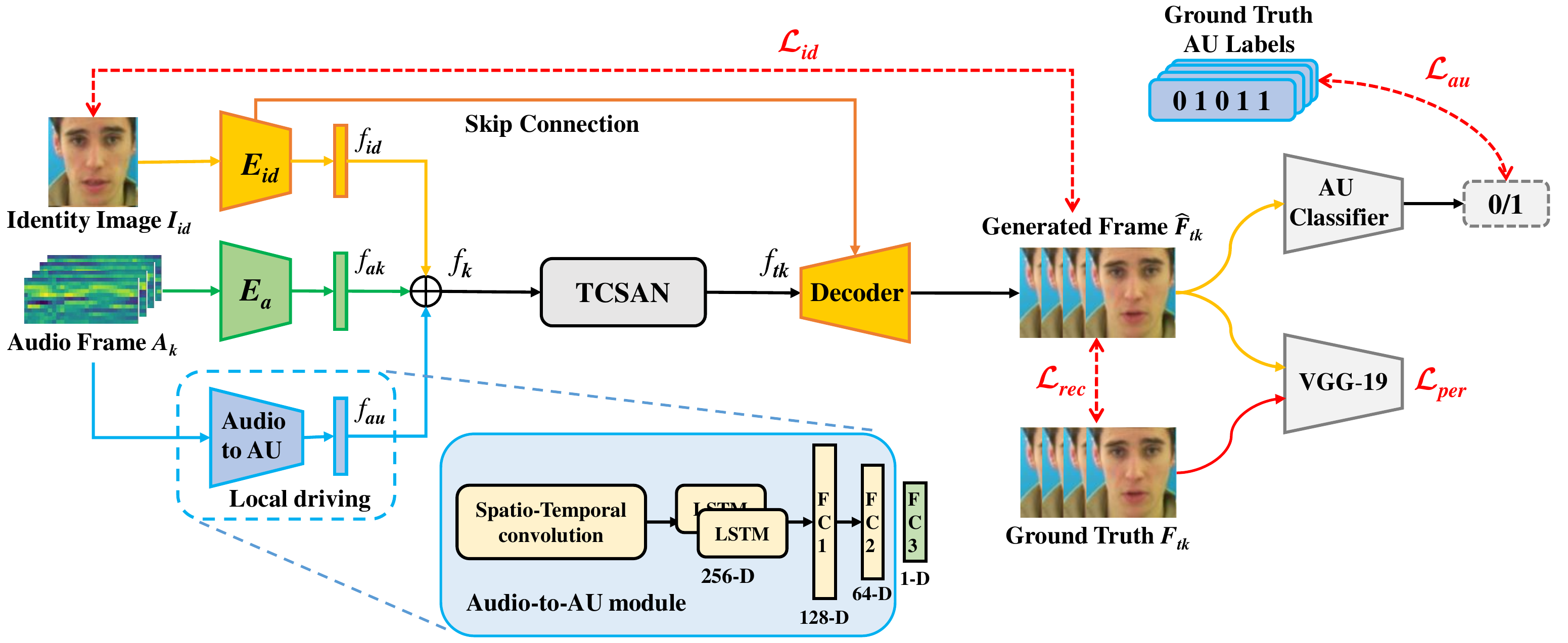}
  \caption{Pipeline of our proposed system. The TCSAN module can effectively integrate multimodal features, and the audio-to-AU module can obtain speech-related facial AUs as local driving information of the mouth movement.}
  \label{fig.2}
\end{figure*}

\section{Proposed Method}

The architecture of the proposed system is illustrated in Fig.~\ref{fig.2}. The generator contains an identity encoder, audio encoder, audio-to-AU module, TCSAN module, and image decoder. In addition, we propose AU classifier to supervise whether the AU information of the generated image is accurate. We use the VGG-19 network \cite{simonyan2014very} to extract high-level features from the generated frame and ground truth frame and compare the differences between them. In the following section, we describe each module in detail.

\subsection{Network architecture}\label{3.1}

\textbf{Identity Encoder:} The identity encoder \(E_{id}\) contains four 2D convolution layers and a fully connected layer. Each convolution layer is followed by a ReLU activation function. The input \(I_{id}\) is a face image resized to 112 \(\times\) 112 as an identity image. This can be a frame randomly selected from a video. For convenience, we used the first frame of each video in this study. The output of \(E_{id}\) is a 512-dimensional identity feature vector, \(f_{id}=E_{id}(I_{id})\).

\noindent\textbf{Audio Encoder:} The audio encoder \(E_{a}\) contains five 2D convolution layers and two fully connected layers; each convolution layer is followed by batch normalization and ReLU activation function. We preprocessed the audio sequence before input. Specifically, we extracted mel-frequency cepstral coefficient (MFCC) features, and then used a fixed-size sliding window to crop MFCC segments continuously. Finally, the inputs are continuous MFCC frames \(A=[A_{1}, A_{2}, ..., A_{k}, ..., A_{n}]\), where the size of \(A_{k}\) is 12\(\times\)28. The outputs are a series of 512-dimensional audio feature vectors \(f_a=[f_{a1}, ..., f_{ak}, ...,  f_{an}]\), where \(f_{ak}=E_{a}(A_{k})\).

\begin{figure}[!t]
  \centering
  \includegraphics[width=0.6\linewidth]{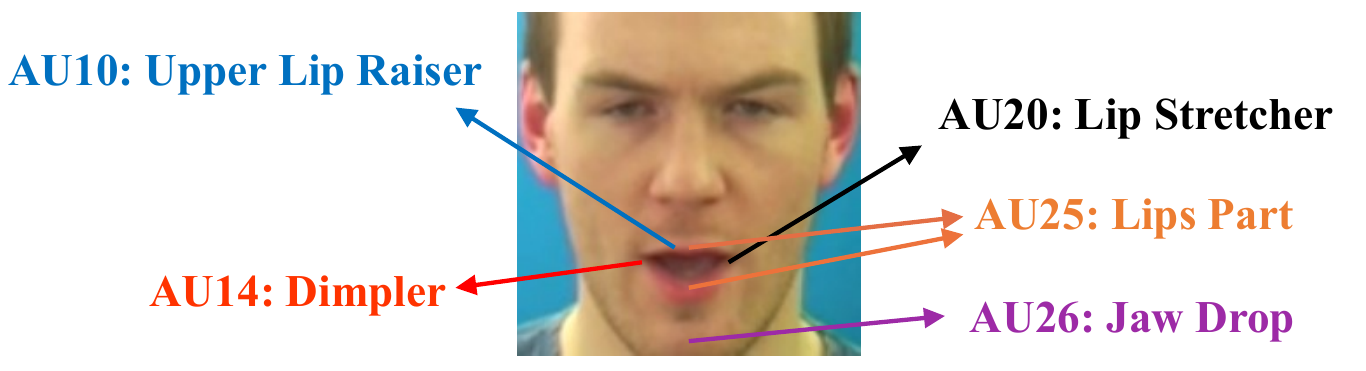}
  \caption{Interpretation of speech-related AUs}
  \label{fig.3}
\end{figure}

\noindent\textbf{Audio-to-AU Module:} To guide the muscle movements in the mouth region more accurately, we propose an audio-to-AU module to extract speech-related AU information from the speech in real-time. It was pre-trained with paired audio and AU data. The pre-trained audio-to-AU module is shown in Fig.~\ref{fig.2}. In the pre-training stage of the audio-to-AU module, the input MFCC features are convolved in the frequency domain and time domain. They then pass through two LSTM layers and three fully connected layers, and the sigmoid activation function is used to obtain the probability of AU occurrence. After pre-training, we removed the last fully connected layer of this network as the audio-to-AU module and added it to our framework, because the multidimensional AU representation obtained by the penultimate fully connected layer is more conducive to the model learning AU information than a one-dimensional label. Because talking is a facial movement driven by multiple facial muscles in the mouth region, multiple audio-to-AU modules are necessary to extract this local facial information. 
Based on anatomy knowledge \cite{martinez2017automatic}, five speech-related AUs were selected in this work, namely AU10, AU14, AU20, AU25, and AU26. The detailed interpretation is presented in Fig.~\ref{fig.3}. Each output of the audio-to-AU module is a 64-dimensional AU feature vector \(f_{au_i}\), where \(i\) is the number of the \(i\)th AU. We concatenate the five 64-dimensional vectors and finally obtain the 320-dimensional AU feature vector \(f_{au}=f_{au_{10}}\oplus f_{au_{14}}\oplus f_{au_{20}}\oplus f_{au_{25}}\oplus f_{au_{26}}\) .

\begin{figure}[!t]
  \centering
  \includegraphics[width=0.75\linewidth]{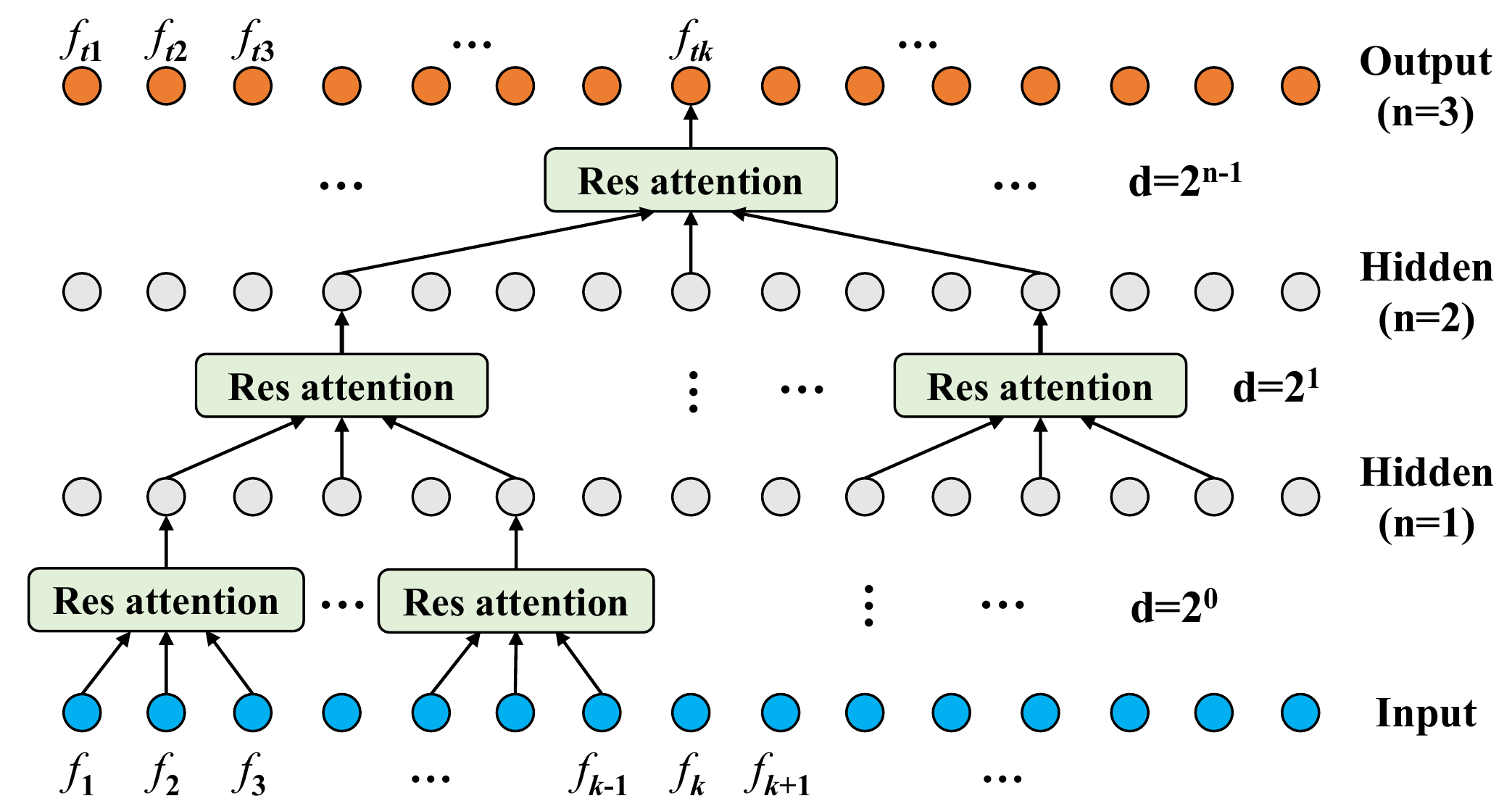}
  \caption{Proposed dilated non-causal temporal convolutional self-attention network. "Res attention" means the residual attention block.}
  \label{fig.4}
\end{figure}

\begin{figure}[!t]
  \centering
  \includegraphics[width=0.8\linewidth]{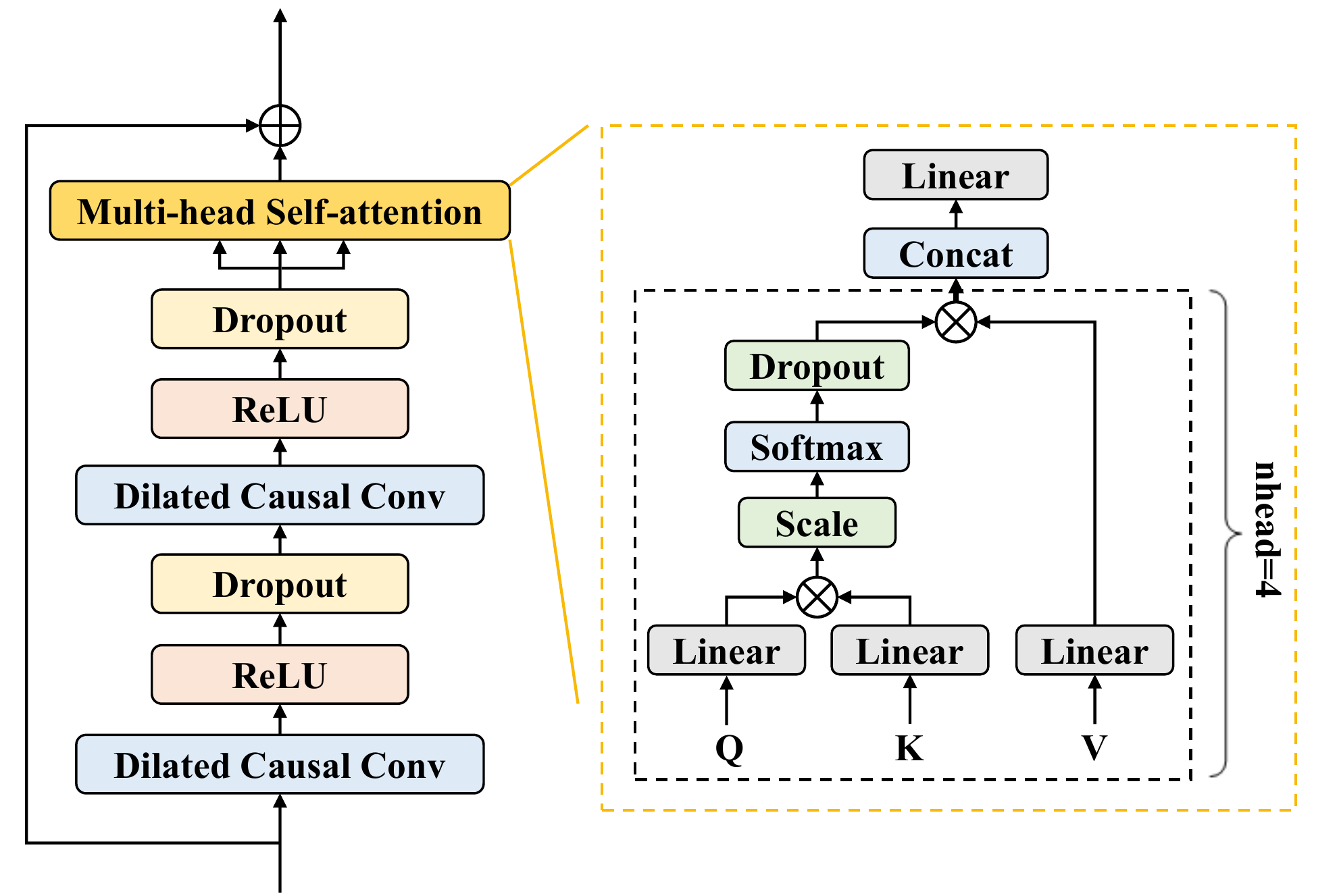}
  \caption{Residual attention block in TCSAN. In the multi-head self-attention module, Q = K = V.}
  \label{fig.5}
\end{figure}

\noindent\textbf{Temporal Convolutional Self-Attention Network:} We introduce the TCSAN module for sequence modeling to replace the RNN module in our earlier conference version \cite{chen2021talking}. The TCSAN module can simultaneously capture longer sequence information and learn the interrelationship among multimodal information. The causal TCN proposed by Bai et al. \cite{bai2018empirical} was only designed to use previously observed information. However, the current state of mouth movement is not only affected by the past timing information, but also depends on future information. As shown in Fig.~\ref{fig.4}, we use a non-causal TCN instead of a causal TCN, which can also capture the information in the next few frames. Dilated convolution is employed to enable an exponentially large receptive field. The operation \(T\) of the dilated convolution on element \(s\) of the sequence \(X\) is defined as:

\begin{equation}
  \ {T(s)} = (X*_{d}F)(s)= \sum_{i=0}^{k-1} F(i)\cdot X_{s+d\cdot (\frac{k-1}{2}- i)},
  \label{eq1}
\end{equation}

where \(F\) is the filter, \(k\) is the filter size (\(k\)= 3, 5, 7, ...), \(d\) is the dilation factor, and \(\frac{k-1}{2}- i\) represents that the number of elements before and after \(s\) are symmetric. The size of \(d\) is fixed in the same hidden layer, and there are multiple differences between adjacent layers. It increases exponentially with the network depth, where  \(d=2^{n-1}\), \(n\) is the \(n\)th hidden layer. Although the receptive field is effectively expanded, the relationship learning of the cross-modal features has not been solved. Thus, we add multi-head self-attention into the TCN to promote the fusion of cross-modal features. The residual attention block in TCSAN is shown in Fig.~\ref{fig.5}. In contrast to other works that weight each frame in the sequence, we use multi-head self-attention to add weights to different elements of the cross-modal features in the frame.

An attention function can be described as mapping a query and a set of key--value pairs to the output. We use all the keys to calculate the dot products of the query and divide each dot product by \(\sqrt{D_k}\), where \(D_{k}\) represents the dimension of the keys. The softmax function is used to obtain the weights of the values. Then, we can calculate the output of the attention function by Eq.\ref{eq2}:
\begin{equation}
  \ {Att (Q,K,V)} = softmax (\frac{QK^T}{\sqrt{D_k}})V,
  \label{eq2}
\end{equation}
where Q, K, and V represent the matrices of queries, keys, and values, respectively. Instead of performing a single calculation, we project the queries, keys, and values \(h\) times with different learned linear projections to improve performance. We set the number of heads \(h\) to 4 in our experiments. The \(h\) results are concatenated and once again projected. The calculation flow is shown on the right side of Fig.~\ref{fig.5}. Then, we obtain the final output.
\begin{equation}
  \ {M_h(Q,K,V)} = Concat (Att_1,Att_2,...,Att_h)W.
  \label{eq3}
\end{equation}

In Eq.\ref{eq3}, \(W\) is the projection matrix. We set \(Q = K = V = Z_{n}\), where \(Z_{n}\) is the hidden representation in the residual attention block. This attention method is called multi-head self-attention. The \(Z_{n}\) is calculated \(h\) times without sharing parameters, and the h results are projected to \(\hat{Z}_{n}\). Therefore, the multi-head self-attention is defined as Eq.\ref{eq4}:
\begin{equation}
  \ {\hat{Z}_n} = M_h(Z_n).
  \label{eq4}
\end{equation}

In our experimental settings, the filter size \(k\) of TCSAN is fixed at 3. We concatenate the identity feature \(f_{id}\), the audio feature \(f_{ak}\), and the speech-related AUs feature \(f_{au}\) to obtain \(f_{k}=f_{id}\oplus f_{ak}\oplus f_{au}\), where $k$ represents the \(k\)th frame in the sequence. The feature sequence is then input into TCSAN  and we finally obtain \(f_{t}=[f_{t1}, f_{t2}, ..., f_{tk}, ..., f_{tn}]\), where \(f_{tk} = TCSAN(f_k)\).

\noindent\textbf{Image Decoder:} The image decoder is used to generate a talking head video. It consists of a fully connected layer and six transposed convolutional layers. To preserve the input identity information and facial texture, we utilize a structure similar to U-Net \cite{ronneberger2015u}, which uses a skip connection between the identity encoder and the image decoder. From the input feature \(f_{t}=[f_{t1}, f_{t2}, ..., f_{tk}, ..., f_{tn}]\) , we can obtain the decoded image sequence \( \hat{F_{t}}=[\hat{F}_{t1}, \hat{F}_{t2}, ..., \hat{F}_{tk}, ..., \hat{F}_{tn}]\).

\noindent\textbf{AU Classifier:} The AU classifier is used to predict the occurrence probability of the speech-related AUs in the generated frames. It consists of four convolution layers and three fully connected layers. Every two convolution layers are followed by a max-pooling layer. Then, the sigmoid activation function is used to obtain the occurrence probability of each AU. To focus on the mouth region, we input only the lower face of the generated image. Binary cross-entropy loss is used to calculate the loss between the predicted and ground truth AU labels \cite{shao2018deep}.
\begin{equation}
  \ \mathcal{L}_{bce} = - \frac{1}{n_{au}}\sum_{i=1}^{n_{au}} w_i[y_i\log \hat{y}_i+(1-y_i )\log(1-\hat{y}_i)],
  \label{eq.5}
\end{equation}
where \(y_i\) represents the ground truth label of the \(i\)th AU, which is 1 if the AU is occurrence and 0 otherwise, and \(\hat{y}_i\) represents the corresponding predicted probability of occurrence. The occurrence rates of AUs are imbalanced for most facial AU detection benchmarks \cite{martinez2017automatic}. Because AUs are not independent of each other, an imbalance in the training data negatively impacts performance. Therefore, we add weight \(w_i\) in Eq.\ref{eq.5} to alleviate the data imbalance problem, where \(w_i=\frac{(1/r_i)n_{au}}{\sum_{i=1}^{n_{au}}{1/r_i}}\), \(r_i\) is the occurrence rate of the \(i\)th AU in the training set.

Some AUs rarely appear in the training set; therefore, network prediction tends to be absent. To alleviate this problem, we introduce a weighted multi-label Dice coefficient loss \cite{milletari2016v}. 
\begin{equation}
  \ \mathcal{L}_{dice} = \frac{1}{n_{au}}\sum_{i=1}^{n_{au}} w_i[1-\frac{2y_i\hat{y}_i+\epsilon}{y_i^2+\hat{y}_i^2+\epsilon}],
  \label{eq6}
\end{equation}
where \(\varepsilon\) is a smooth term. Thus, our AU loss is defined as

\begin{equation}
  \ \mathcal{L}_{au} = \mathcal{L}_{bce} +\mathcal{L}_{dice}. 
  \label{eq7}
\end{equation}

\begin{figure*}[!t]
  \centering
  \includegraphics[width=0.9\linewidth]{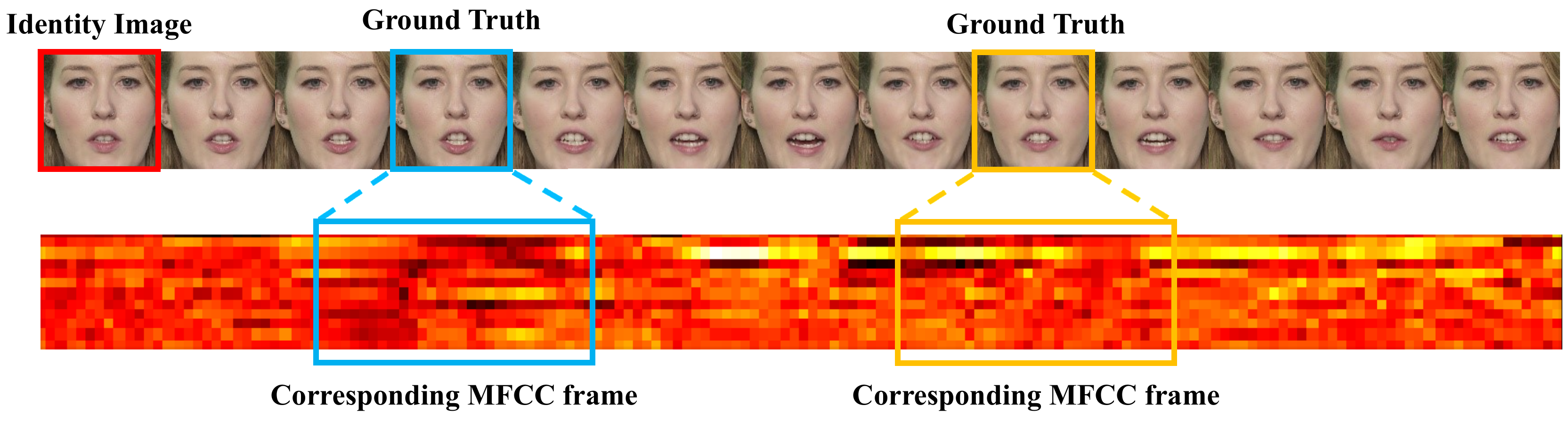}
  \caption{Input details of video stream and audio stream in the model.}
  \label{fig.6}
\end{figure*}

\noindent\textbf{VGG-19 Network:} In order to improve the quality of generated images, we use perceptual loss \cite{johnson2016perceptual} to reflect perceptual-level similarity of images. The pre-trained VGG-19 network \cite{simonyan2014very} was adopted as the perceptual feature extractor. The perceptual loss is defined as
\begin{equation}
  \ \mathcal{L}_{per}(F_{tk},\hat{F}_{tk}) = \frac{1}{n}\sum_{i=1}^n ||\phi_i(F_{tk}) - \phi_i(\hat F_{tk}) ||_1,
  \label{eq8}
\end{equation}
where \(\phi_i\) denotes the \(i\)th feature extraction layer of the VGG-19 network. 

\subsection{Loss functions}

In addition to the aforementioned AU loss and perceptual loss, reconstruction loss and identity loss are also considered. The reconstruction loss \({L}_{rec}\) is used to minimize the pixel-level difference between the generated image \(\hat{F}_{tk}\) and the real image \(F_{tk}\).
\begin{equation}
  \ \mathcal{L}_{rec}(\hat{F}_{tk}, F_{tk}) =  ||\hat F_{tk} - F_{tk} ||_1.
  \label{eq9}
\end{equation}

The identity loss \({L}_{id}\) is used to maintain the identity information and reduce the jitter effect of the generated video. Specifically, it penalizes the difference between the upper face of \(\hat{F}_{tk}\) and the upper face of \(I_{id}\).

\begin{equation}
  \ \mathcal{L}_{id}(\hat F_{tk_p}, I_{id_p})=\frac{1}{W\times\frac{H}{2}}  \sum_{p\in [0,W]\times[0,\frac{H}{2}]}||\hat F_{tk_p} - I_{id_p} ||_1,
  \label{eq10}
\end{equation}
where $W$ and $H$ are the width and height of each frame, respectively; \(\hat{F}_{tk_p}\) is the upper half of the generated image \(\hat{F}_{tk}\), and \(I_{id_p}\) is the upper half of the input image \(I_{id}\).

Finally, the overall loss of the proposed framework is defined by Eq. \ref{eq11}.
\begin{equation}
  \ \mathcal{L}_{total} =\lambda_{rec} \mathcal{L}_{rec} +\lambda_{id} \mathcal{L}_{id} +\lambda_{per}\mathcal{L}_{per} +\lambda_{au}\mathcal{L}_{au},
  \label{eq11}
\end{equation}
where \(\lambda_{rec}\), \(\lambda_{id}\), \(\lambda_{per}\), and \(\lambda_{au}\) are the trade-off parameters.

\section{Experimental Setup}

\subsection{Dataset}

We conducted extensive experiments on the GRID \cite{cooke2006audio} and TCD-TIMIT datasets \cite{harte2015tcd}. The GRID dataset is a large audio--visual corpus, which consists of high-quality audio and video recordings of 1000 sentences spoken by each of 33 speakers. The TCD-TIMIT dataset has high-quality audio and video footage of 59 speakers, each uttering approximately 100 phonetically rich sentences. Because our model generates frontal talking head, and the above two datasets are high-quality frontal talking head collected in the laboratory environment, they are very suitable for our model. In our experiments, the GRID dataset was divided into a training set and a testing set at a ratio of 8: 2, with 27 speakers in the training set and six speakers in the test set. For the TCD-TIMIT dataset, we assigned the audio and video footage of 49 speakers as the training set and that of 10 speakers as the testing set.

As shown in Fig.~\ref{fig.6}, each recording in the dataset was processed into a video stream and an audio stream. For the video stream, we extracted all the video frames and used the Dlib toolkit \cite{king2009dlib} of the HOG-based face detection algorithm to detect all the faces. Then, we cropped the face regions and resized them to 112\(\times\)112. Similar to existing works\cite{wu2020cascade,guo2020deep,tang2020fine}, we used OpenFace \cite{baltrusaitis2018openface,baltruvsaitis2015cross} to detect the AU labels of each face. For the audio stream, we extracted 12-dimensional mel-scale frequency cepstral coefficients (MFCC) features (excluding energy dimension). We used the first frame of each video as input identity information. For audio input, we tried different lengths of audio frames and found that 280 ms performed best. Then, we aligned the middle of each audio frame with a corresponding video frame. The audio sliding window is defined to slide synchronously with the video frame. We compared the images generated by each audio frame with the corresponding ground truth. 

\subsection{Training details}

Our full network was implemented using Pytorch and trained for 20 epochs on a single NVIDIA Titan V GPU. The audio-to-AU module and the AU classifier were pre-trained on the GRID dataset, and when experimenting on the TCD-TIMIT dataset, they were refined to fit the new dataset. The VGG-19 network was pre-trained on the ImageNet dataset \cite{russakovsky2015imagenet}. After pre-training of the above modules, we added them to the full model. We adopted the Adam optimizer with \(\beta_1=0.5\) and a learning rate of 0.0002 during the full model training. The weights of  \(\lambda_{rec}\), \(\lambda_{id}\), \(\lambda_{per}\), and \(\lambda_{au}\) are 1.5, 1.5, 0.07, and 0.02, respectively. We used \(\mathcal{L}_{au}\) to fine-tune the audio-to-AU and AU classifiers during the full model training, and adopt the Adam optimizer with learning rates of 1e-6 and 1e-7, respectively. The parameters of the VGG-19 network were frozen during the training of the full model.

\section{Results}
\subsection{AU Detection with the AU Classifier}

To enable the AU classifier to judge whether the generated image contained the correct AU information, we used the GRID dataset to pre-train it. The training data were images and the corresponding speech-related AU labels. Openface \cite{baltrusaitis2018openface} was used to extract the speech-related AU labels of each image. 

The results of our pre-trained AU classifier on the GRID dataset are listed in Table \ref{tab:1}. The average F1 score of AUs was 0.945, and the average accuracy of AUs was 95.33\% on the training set. This proves that our AU classifier can effectively capture the AU information. On the testing set, it achieved an average F1 score of 0.871 and an average accuracy of 89.62\%, which shows that our pre-trained AU classifier has a strong generalization ability.

\begin{table}[!t]
\caption{Results of our pre-trained AU classifier on real images of the GRID training set and testing set.}
\begin{center}
\small
\setlength\tabcolsep{3pt}
\begin{tabular}{ccccc}
\toprule
\multirow{2}*{AU Number}&\multicolumn{2}{c}{Training Set}&\multicolumn{2}{c}{Test Set}\\
\cline{2-5}
&F1 score& Accuracy& F1 score& Accuracy \\
\midrule
AU10 &0.971 &96.81\%&0.875 &88.60\% \\
AU14 &0.896&95.54\%&0.752 &91.07\%\\
AU20  &0.955&95.13\%&0.896 &89.85\%\\
AU25  &0.983&97.16\%&0.979 &96.38\%\\
AU26  &0.920&91.99\%&0.854 &82.18\%\\
\midrule
Average  &0.945&95.33\%&0.871 &89.62\%\\
\bottomrule
\end{tabular}
\end{center}
\label{tab:1}
\end{table}

To further verify the effectiveness of the AU classifier, we used it to detect the speech-related AUs of the images generated by our proposed method. The detection results are listed in Table~\ref{tab:2}. Because the ground truth AU labels were extracted by Openface \cite{baltrusaitis2018openface}, the Openface detection results were accurate. The AU classifier was pre-trained on the GRID dataset; hence, when experimenting on the TCD-TIMIT dataset, it was necessary to use the new dataset to refine it to adapt to the new domain. The detection results of the AU classifier on the GRID test set were very close to those of OpenFace. On the TCD-TIMIT dataset, the AU classifier detection results on average F1 score were lower than those of Openface detection; however, the accuracy was higher. The main reason for this is that the distribution of AUs on the TCD-TIMIT dataset is unbalanced; hence, the prediction of unbalanced AU tends to be absent, resulting in lower F1 score and higher accuracy. In general, the AU classifier can determine whether the generated image contains the correct AU information.

\begin{table}[!t]
\caption{AU detection results of the images generated by our proposed method on AU classifier and Openface, respectively. Because the ground truth AU labels of the real images were extracted using OpenFace, the Openface detection results were accurate.}
\begin{center}
\small
\setlength\tabcolsep{3pt}
\begin{tabular}{ccccc}
\toprule
\multirow{2}*{Method}&\multicolumn{2}{c}{GRID}&\multicolumn{2}{c}{TCD-TIMIT}\\
\cline{2-5}
&Avg. F1&Avg. Acc. &Avg. F1& Avg. Acc. \\
\midrule
AU classifier  &0.745 &79.43\%&0.500 &89.26\%\\
Openface \cite{baltrusaitis2018openface} &0.751 &80.56\%&0.590 &86.51\% \\
\bottomrule
\end{tabular}
\end{center}
\label{tab:2}
\end{table}

\subsection{Effectiveness of Temporal Convolutional Self-Attention Network}

To evaluate the quality of the generated images, we adopted the reconstruction metrics peak signal to noise ratio (PSNR) and structural similarity index measure (SSIM) \cite{wang2004image}. For the lip-sync performance, we verified the recognition accuracy and F1 score of the five selected speech-related AUs. Specifically, we used the OpenFace toolkit \cite{baltrusaitis2018openface,baltruvsaitis2015cross} to detect the state of the five selected AUs (activated or not) in each generated frame, and then compared them with ground truth labels. 

We first compared different cross-modal representation fusion methods without adding the Audio2AU module. Table \ref{tab:3} shows the quantitative results with different representation fusion methods on the GRID and TCD-TIMIT test sets. It may be observed that TCN performed better than GRU in both image quality and lip-sync accuracy on the two datasets, which shows that TCN is better at capturing temporal relationships. To promote cross-modal representation fusion, we used multi-head self-attention and multi-head co-attention respectively before the TCN module. Because of the relationship learning of cross-modal features, compared with TCN, the image quality and lip-sync accuracy had a small improvement on the GRID dataset and and a relatively large improvement on the TCD-TIMIT dataset. After we replaced other fusion methods with TCSAN, all metrics were further improved. Especially on the TCD-TIMIT dataset, compared with using multi-head co-attention before the TCN module, PSNR improved by 0.19, while average F1 score improved by 1.5\%. This demonstrates that our proposed TCSAN has obvious advantages in both temporal modeling and relationship learning of cross-modal representations.

\begin{table*}[!t]
\caption{Quantitative results with different representation fusion methods on the GRID test set and TCD-TIMIT test set. Mh-att is the multi-head self-attention module and Co-att is the multi-head co-attention module. Avg. F1 and Avg. Acc. are the average F1 score and average accuracy (\%) of speech-related AUs, respectively.}
\begin{center}
\footnotesize
\setlength\tabcolsep{3pt}
\scalebox{0.96}{
\begin{tabular}{lcccccccc}
\toprule
\multirow{2}*{Method}&\multicolumn{4}{c}{GRID}&\multicolumn{4}{c}{TCD-TIMIT}\\
\cline{2-9}
&PSNR↑&SSIM↑&Avg. F1&Avg. Acc.&PSNR↑&SSIM↑& Avg. F1&Avg. Acc. \\
\midrule
GRU & 29.403 & 0.753&0.735 &79.27
&26.132 & 0.741&0.550 &82.54 \\
TCN & 29.671 &0.765 &0.739&79.45
&26.290 & 0.747 & 0.553 &83.24\\
Mh-att+TCN & 29.757 &0.766 &0.740&79.55
&26.377 & 0.750&0.561 &83.82\\
Co-att+TCN & 29.720 &0.765 &0.740&79.50
&26.319 & 0.753&0.563 &84.83\\
TCSAN (ours) & \textbf{29.817} &\textbf{0.767}&\textbf{0.744}&\textbf{80.01}
&\textbf{26.509} & \textbf{0.757}&\textbf{0.578} &\textbf{85.25}\\
\bottomrule
\end{tabular}
}
\end{center}
\label{tab:3}
\end{table*}

\begin{figure*}[!t]
  \centering
  \includegraphics[width=0.95\linewidth]{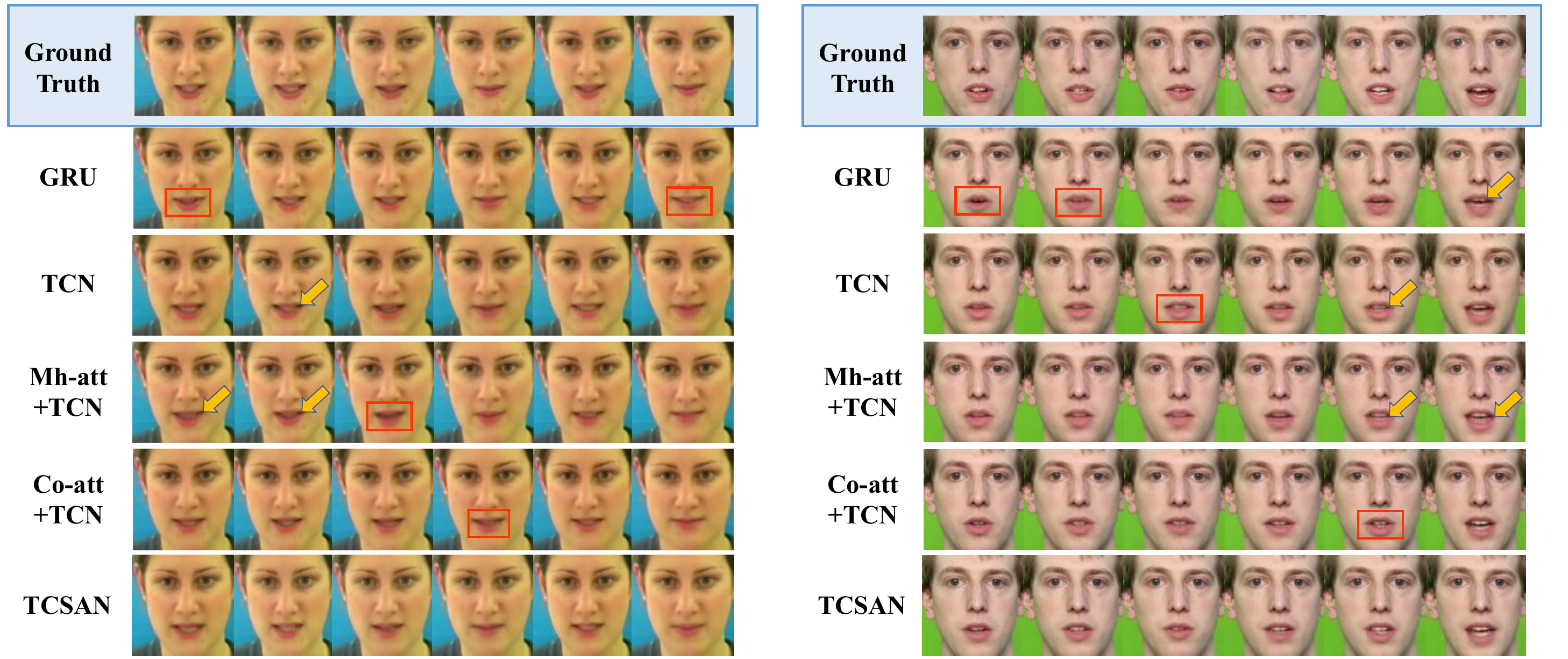}
  \caption{Qualitative results with different representation fusion methods on the GRID test set and TCD-TIMIT test set. Mh-att is the multi-head self-attention and Co-att the multi-head co-attention. Each row represents a sequence of talking faces. The arrows indicate the examples of blurring, and the red boxes indicate examples of mouth movement out of sync.}
  \label{fig.7}
\end{figure*}

To further demonstrate the effectiveness of our TCSAN module, we show the qualitative results of the GRID test set and TCD-TIMIT test set in Fig.~\ref{fig.7}. Two different audio sequences were used to generate the talking faces for two speakers. Compared with other cross-modal representation fusion methods, our TCSAN performed better in terms of both image quality and lip-sync accuracy. For example, the facial details, such as the teeth indicated by the arrows in the figure, generated by other methods are blurry. Red boxes are used to mark images where the lips are out of sync. In addition, the image generated by our TCSAN is better than that by the other methods in terms of the change in mouth motion. It is intuitively proved that our TCSAN can better learn the temporal relationship of the cross-modal representations.

\subsection{Effectiveness of Audio-to-AU Module}

To evaluate the impact of our Audio2AU module, Table \ref{tab:4} shows the quantitative results with and without using the Audio2AU module on the two datasets. It can be clearly noted that after adding Audio2AU, all the metrics were further improved. The results show that the Audio2AU module not only had a positive impact on improving the lip-sync accuracy, but also potentially improved the image quality. Similar to our earlier conference version \cite{chen2021talking}, we also attempted to use the GAN training method \cite{goodfellow2014generative} in our full model. Unfortunately, the image quality and lip-sync accuracy did not improve. We attribute this to the difficulty of training the GAN, or to our model already having sufficient penalties to supervise the training process.

\begin{table*}[!t]
\caption{Quantitative results with/without Audio2AU module on the GRID test set and TCD-TIMIT test set. Avg. F1 and Avg. Acc. are the average F1 score and average accuracy (\%) of speech-related AUs, respectively. We also attempted to use a GAN for comparison.}
\begin{center}
\scriptsize
\scalebox{0.84}{
\begin{tabular}{lcccccccc}
\toprule
\multirow{2}*{Method}&\multicolumn{4}{c}{GRID}&\multicolumn{4}{c}{TCD-TIMIT}\\
\cline{2-9}
&PSNR↑&SSIM↑&Avg. F1&Avg. Acc.&PSNR↑&SSIM↑& Avg. F1&Avg. Acc. \\
\midrule
TCSAN & 29.817 & 0.767&0.744 &80.01
&26.509 & 0.757&0.578 &85.25 \\
Audio2AU & 29.838 & 0.769 & \textbf{0.751} &80.92
&26.201 & 0.745&\textbf{0.590} &84.92 \\
TCSAN+Audio2AU & \textbf{29.971} &\textbf{0.772}&\textbf{0.751}&\textbf{81.15}
&\textbf{26.601} & \textbf{0.760}&\textbf{0.590} &\textbf{86.51}\\
\bottomrule
\end{tabular}
}
\end{center}
\label{tab:4}
\end{table*}

\begin{figure*}[!t]
  \centering
  \includegraphics[width=\linewidth]{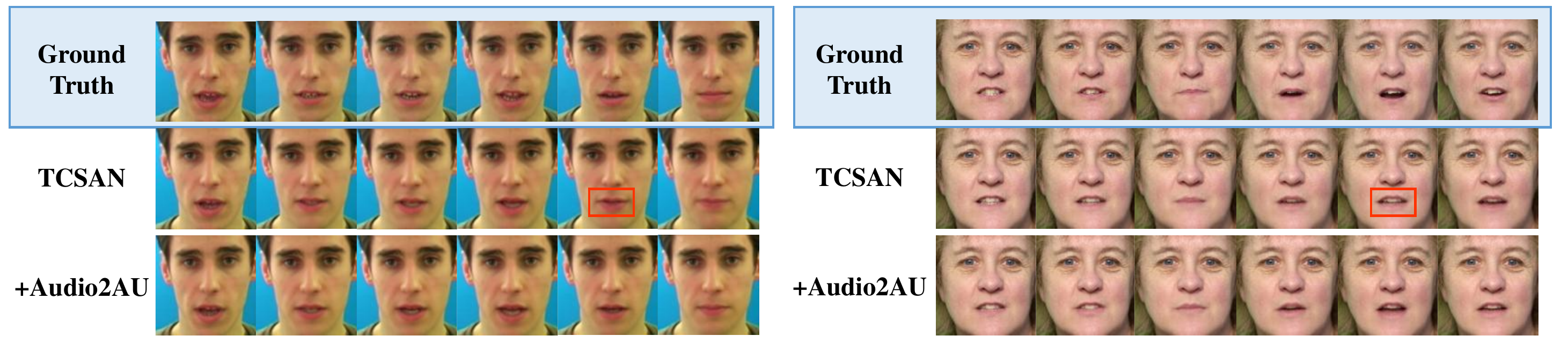}
  \caption{Qualitative results with/without Audio2AU module on the GRID test set and TCD-TIMIT test set. Each row represents a sequence of talking heads. The red boxes indicate examples of mouth movement out of sync.}
  \label{fig.8}
\end{figure*}

In Fig.~\ref{fig.8}, we show the visualized results using the Audio2AU module. Each row represents a sequence of talking faces. The images generated by the model without Audio2AU were worse than those generated by the model using Audio2AU. We used red boxes to mark poor mouth movements. The model with GAN also performed poorly in terms of lip-sync accuracy and mouth movement. In comparison, the model using TCSAN and Audio2AU performed the best overall.

\subsection{Comparison of the Proposed Model with Other Methods}

\begin{table*}[!t]
\caption{Quantitative results compared with song et al. \cite{song2018talking} and Jamaludin et al. \cite{jamaludin2019you} on the GRID test set and TCD-TIMIT test set. }
\begin{center}
\small
\scalebox{0.8}{
\begin{tabular}{lcccccccc}
\toprule
\multirow{2}*{Method}&\multicolumn{4}{c}{GRID}&\multicolumn{4}{c}{TCD-TIMIT}\\
\cline{2-9}
&PSNR↑&SSIM↑&Avg. F1&Avg. Acc.&PSNR↑&SSIM↑& Avg. F1&Avg. Acc. \\
\midrule
CRAN \cite{song2018talking}& 28.041 & 0.694&0.710 &78.71
&24.381 & 0.650&0.562 &81.41 \\
Speech2Vid \cite{jamaludin2019you}& 28.863 & 0.755&0.738 &78.97
&25.132 & 0.685&0.556 &81.91 \\
Baseline ($\mathcal{L}_{rec}$)  & 28.681 &0.746 &0.725&77.77
&25.003 & 0.717&0.545 &80.96\\
Proposed method & \textbf{29.971} &\textbf{0.772}&\textbf{0.751}&\textbf{81.15}
&\textbf{26.601} & \textbf{0.760}&\textbf{0.590} &\textbf{86.51}\\
\bottomrule
\end{tabular}
}
\end{center}
\label{tab:5}
\end{table*}

To demonstrate the superiority of our method, we compared our model with other methods. We implemented the methods proposed by Song et al. \cite{song2018talking} and Jamaludin et al. \cite{jamaludin2019you} under the same conditions, and used the same training--testing data split as our proposed method. Our baseline uses only reconstruction loss. From Table \ref{tab:5}, we can see that our proposed method exhibited significant improvements in image quality and lip-sync accuracy on the GRID and TCD-TIMIT datasets. Compared with Song et al. \cite{song2018talking}, our proposed method improved the PSNR by 1.93, and the average F1 score of AUs by 4.1\% on the GRID dataset. Our method also improved the PSNR by approximately 2.2 and the average accuracy of AUs by 5.1\% on the TCD-TIMIT dataset. Similarly, our method performed significantly better than that of Jamaludin et al. \cite{jamaludin2019you} on all metrics.

\begin{figure*}[!t]
\centering
  \includegraphics[width=0.98\linewidth]{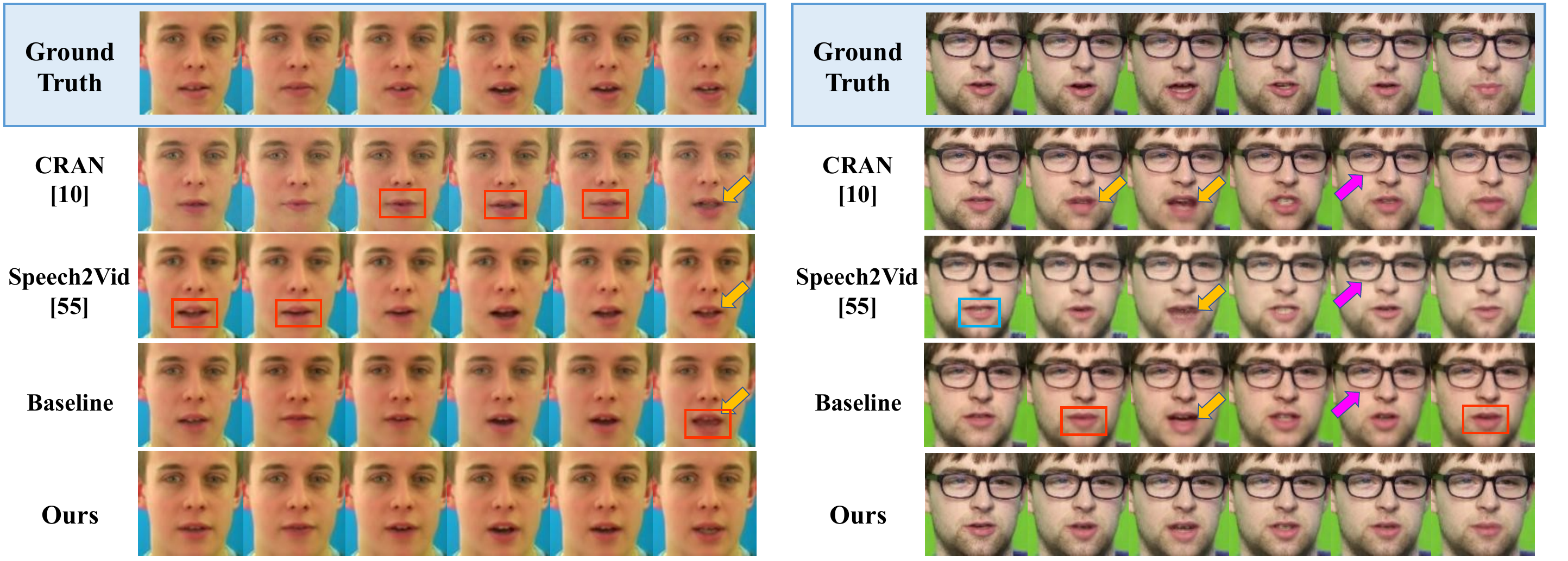}
  \caption{Qualitative results produced by our proposed method and other methods on the GRID test set and TCD-TIMIT test set. We use yellow arrows to indicate the subtle artifacts in the mouth. The pink arrows indicate that the glasses are blurry. The blue box indicate that the teeth are closed, and the red boxes indicate examples of mouth movement out of sync.}
  \label{fig.9}
\end{figure*}

The qualitative results produced by the different methods are also presented in Fig.~\ref{fig.9}. The face images generated by our method are obviously clearer than others. We again use the red boxes to mark the examples of mouth movement out of sync, and use arrows to indicate the examples of blurring, such as teeth and lips. In the picture on the right, the teeth generated by the method of Jamaludin et al. \cite{jamaludin2019you} were closed, which are marked with a blue box. The mouth generated by the method of Song et al. \cite{song2018talking} has subtle artifacts, as indicated by the yellow arrows. The pink arrows indicate that the glasses generated by the other methods are blurry. Our proposed method overcame these shortcomings to perform best in terms of both facial texture and mouth movement.

\begin{figure*}[!t]
\centering
  \includegraphics[width=0.96\linewidth]{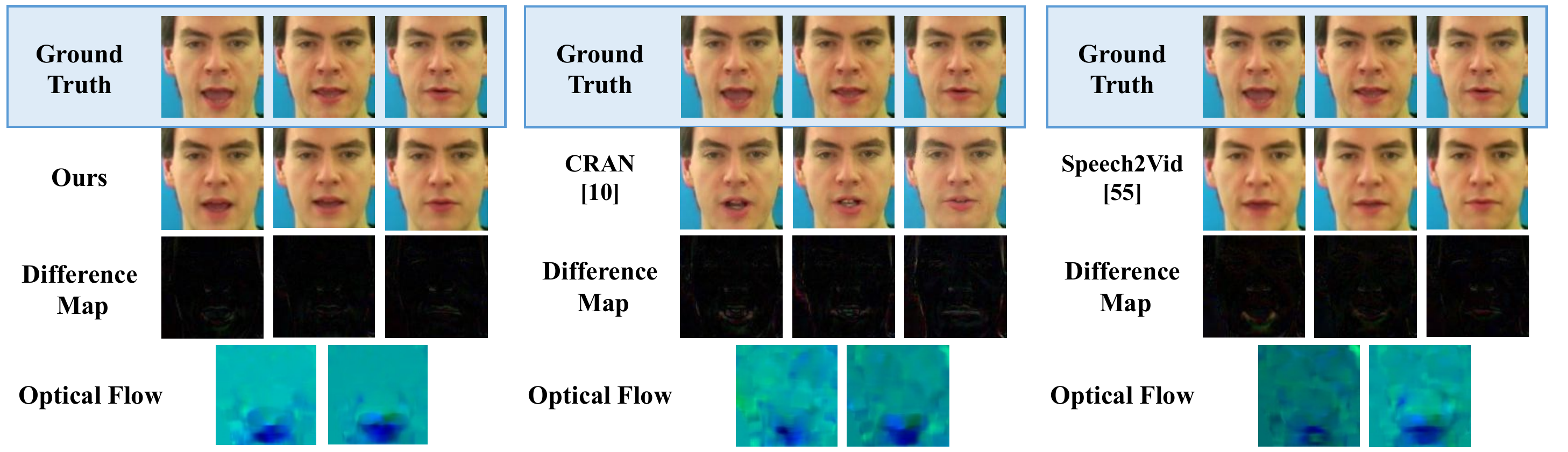}
  \caption{Difference map and optical flow of the face images generated by our proposed method and CRAN \cite{song2018talking} and Speech2Vid \cite{jamaludin2019you}. }
  \label{fig.10}
\end{figure*}

\begin{table*}[!t] 
\caption{Quantitative ablation study on the GRID test set. $\mathcal{L}_{rec}$, $\mathcal{L}_{per}$, $\mathcal{L}_{id}$, and $\mathcal{L}_{au}$ are the reconstruction loss, perceptual loss, identity loss, and AU loss, respectively. Avg. F1 and Avg. Acc. are the average F1 score and average accuracy (\%) of AUs, respectively. }
\begin{center}
 \small
  \centering
   \setlength\tabcolsep{2pt}
   \scalebox{0.68}{
   \begin{tabular}{lcccc cccccccccc}
  \toprule
  \multirow{2}*{Method} &\multirow{2}*{PSNR↑}&\multirow{2}*{SSIM↑}&
    \multirow{1}{*}{Avg.}&\multirow{1}{*}{Avg.}&
                                                                       \multicolumn{2}{c}{AU10}&
                                                                       \multicolumn{2}{c}{AU14}&
                                                                       \multicolumn{2}{c}{AU20}&
                                                                       \multicolumn{2}{c}{AU25}& 
                                                                       \multicolumn{2}{c}{AU26}   \\
                                         
 &&&F1&Acc.& F1& Acc. &F1& Acc.& F1& Acc.& F1& Acc. & F1&Acc. \\
    \midrule
    Baseline ($\mathcal{L}_{rec}$)              & 28.681  & 0.746 &     0.725  & 77.77 &  
    0.745 & 71.14 &  0.477 &82.01 &0.762 &74.86 &0.939 &89.21 &0.703 &71.65\\
    $\mathcal{L}_{rec,per}$           & 28.897  & 0.748   & 0.732  &78.76    &
    0.771 & 75.99 &  0.501 &82.95 &0.742 &74.26 &0.940 &89.10 &0.708&71.50\\
   $\mathcal{L}_{rec, per, id}$            & 29.403 & 0.753  &0.735   &79.27   &
   0.757 & 74.87 &  0.516 &82.20 &0.774 &79.13 &0.943 &89.75 &0.685 &70.41\\
    $\mathcal{L}_{rec, per, id}$ + TCSAN          & 29.817 & 0.767  &   0.744  &80.01&
   0.756 & 74.34 &  0.500 &\textbf{83.03} &0.797 &80.23 &0.946 &90.22 &0.723 &72.21\\
    $\mathcal{L}_{rec, per, id, au}$ + TCSAN     & 29.809 & 0.767  &   0.746  &80.19&
   0.752 & 74.33 &  \textbf{0.517} &82.76 &0.795 &81.17 &\textbf{0.947} &90.34 &0.719 &72.34\\
    Full model(+Audio2AU)    & \textbf{29.971}  & \textbf{0.772}  &\textbf{0.751}  & \textbf{81.15} &
   \textbf{0.781} & \textbf{78.17} &  0.486 &82.93 &\textbf{0.813} &\textbf{81.26} &\textbf{0.947} &\textbf{90.42} &\textbf{0.727} &\textbf{72.98}\\
  \bottomrule
  \end{tabular}
  }
  \end{center}
  \label{tab:6}
\end{table*}

To compare the difference between the ground truth image and the generated image more clearly, we present the difference map in Fig.~\ref{fig.10}. This shows that the mouth regions of the methods of Song et al. \cite{song2018talking} and Jamaludin et al. \cite{jamaludin2019you} were obviously different from the ground truth images. In addition, the edges of the faces generated by Song et al. \cite{song2018talking} also differed from the ground truth images. The difference map of our study shows only a small difference in the mouth. The optical flow is used to represent the motion between the generated frames. The video generated by Song et al. \cite{song2018talking} and Jamaludin et al. \cite{jamaludin2019you} were accompanied by jitter, resulting in chaotic optical flows. The movements of our video frames are concentrated in the mouth; hence, the optical flows are very clear.

\subsection{Ablation study}

To quantify the contribution of each component of the proposed method, we conducted an ablation study on the GRID dataset. As shown in Table~\ref{tab:6}, all metrics were improved after adding perceptual loss and identity loss. In particular, after adding identity loss, PSNR improved significantly. When GRU was replaced with TCSAN, the results were greatly improved. Then, when we added the AU classifier, the lip-sync accuracy had a small improvement, but the PSNR decreased slightly. We believe that when AU loss is used together with the Audio2AU module, it can achieve the maximum effect. Finally, when we added the Audio2AU module, which can improve each other with AU classifier, all the metrics were further improved and achieved the best results. These experiments demonstrate the effectiveness of the proposed method. To observe the results of each AU, we also show the F1 score and accuracy of the five speech-related AUs, respectively, in Table \ref{tab:6}. Most of the results show that our proposed model performed the best. 

\begin{figure}[!t]
\centering
  \includegraphics[width=0.7\linewidth]{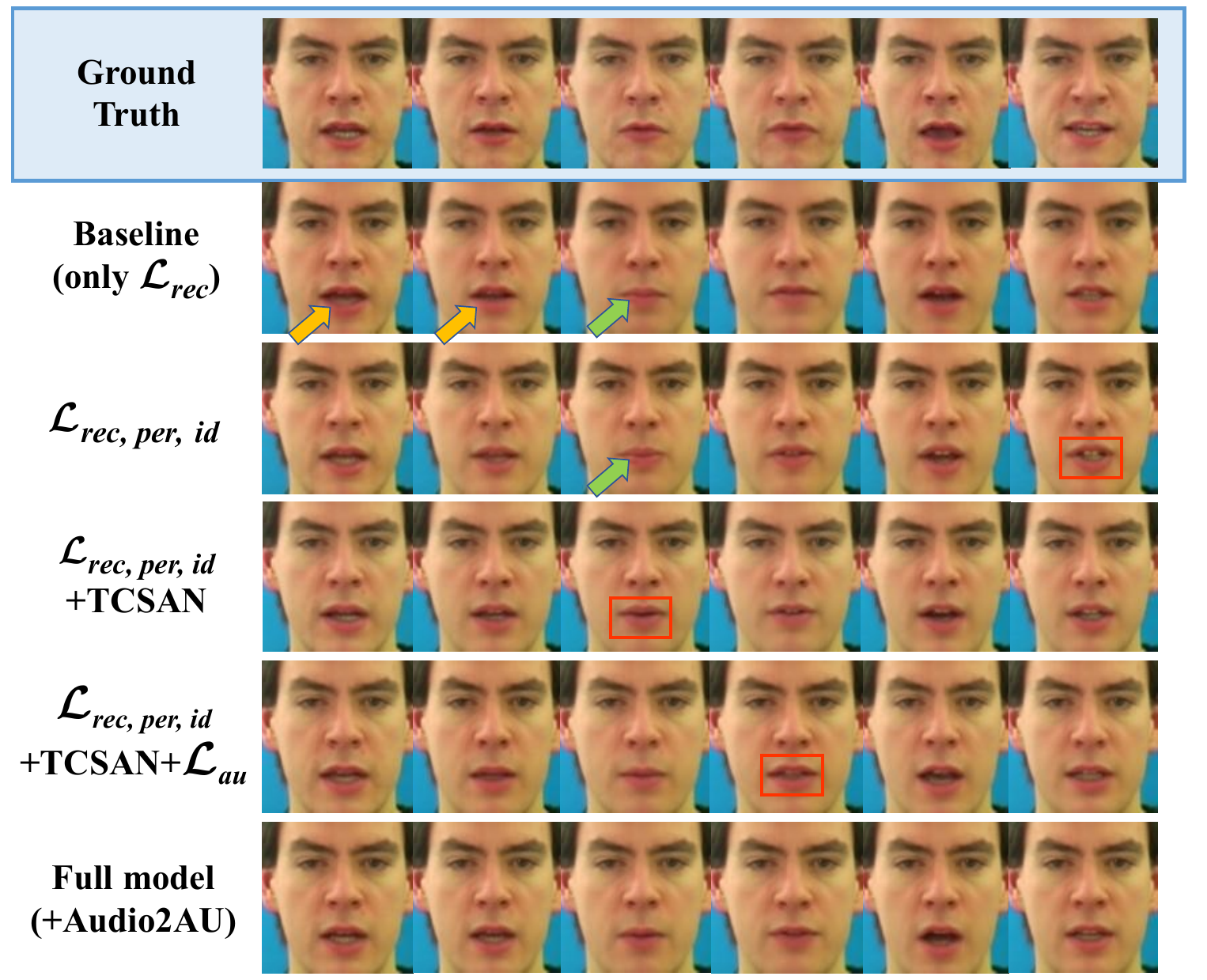}
  \caption{Qualitative results of the ablation study on the GRID test set. Audio2AU refers to our proposed audio-to-AU module. The yellow arrows indicate the teeth are slightly blurred, and the green arrows indicate the lips are slightly lighter. We also use the red boxes to mark the examples of mouth movement out of sync.}
  \label{fig.11}
\end{figure}

In Fig.~\ref{fig.11}, we also show the visualized examples on the GRID test set after adding different components. Red boxes are also used to mark examples where the lips were not very synchronized. Our baseline only uses reconstruction loss, and the generated images were slightly blurred, such as the teeth indicated by the yellow arrows. After adding perceptual loss and identity loss, the image quality, such as facial texture and teeth, was improved. However, the lips indicated by the green arrows were slightly lighter. After adding the TCSAN, the image quality was further refined. After adding AU loss, lip synchronization was improved, but the mouth marked by the red box was open. Finally, the full model with the audio-to-AU module achieved the best results in terms of both image quality and lip-sync accuracy.

\begin{figure}[!t]
\centering
  \includegraphics[width=0.76\linewidth]{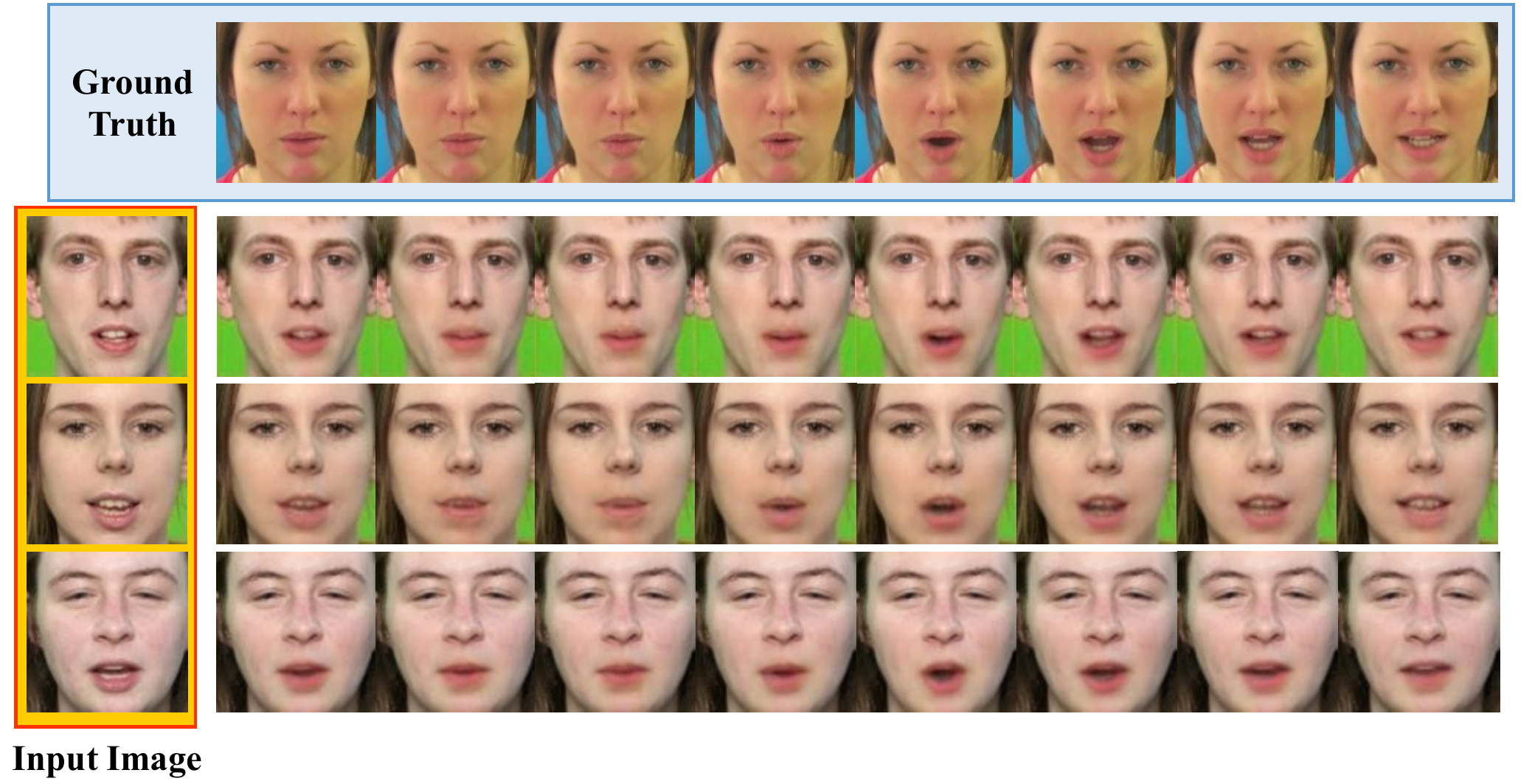}
  \caption{More generated results on the TCD-TIMIT test set using our model pre-trained on the GRID dataset. We use the same audio clip that corresponds to the word "one" for different speakers to generate talking head videos.}
  \label{fig.12}
\end{figure}

\begin{figure}[!t]
\centering
  \includegraphics[width=0.7\linewidth]{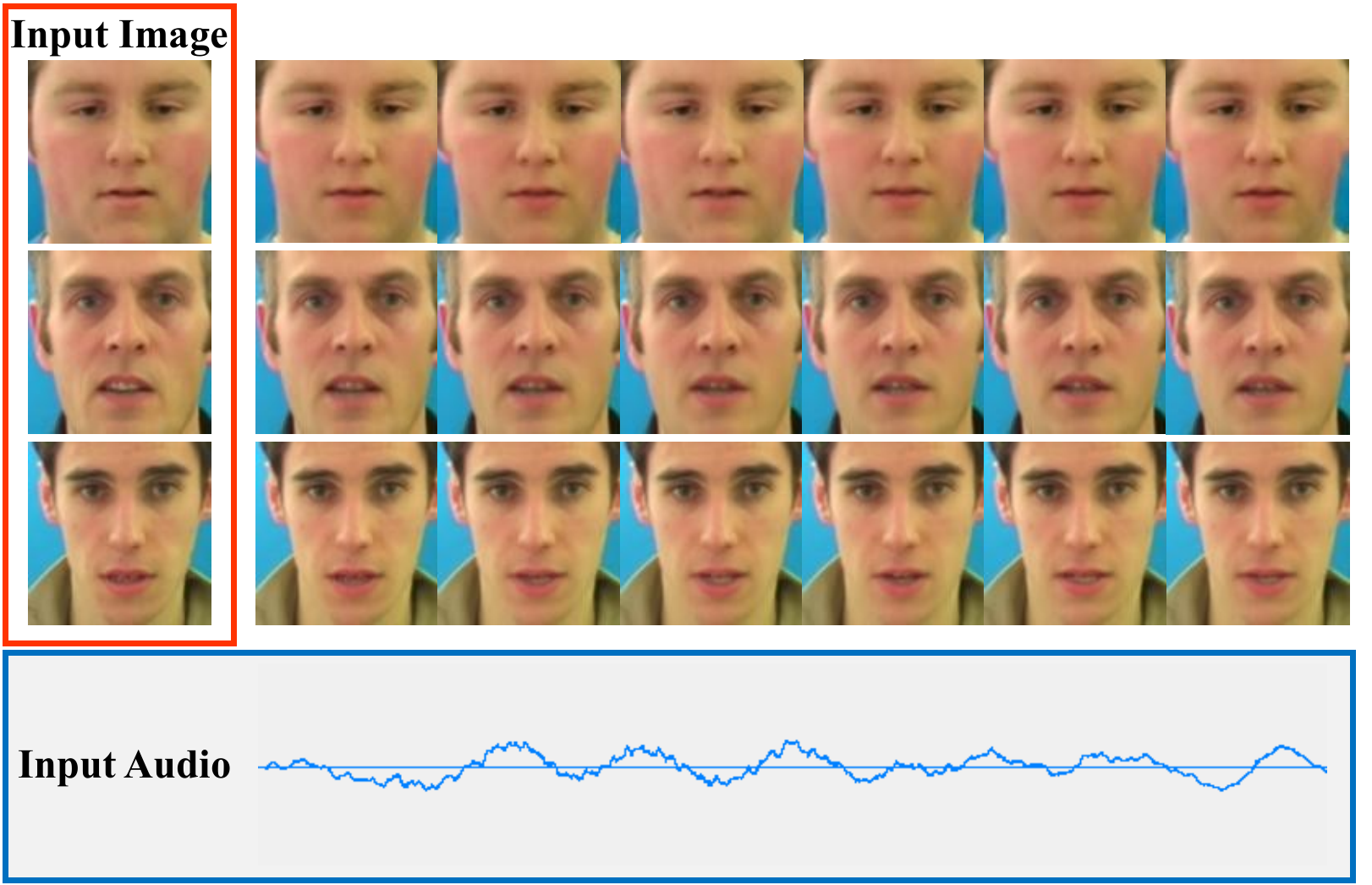}
  \caption{Example of generated videos driven by silent audio with no one speaking and only slight noise. The generated images were consistent with the input image, which obviously suppresses the lip movement.}
  \label{fig.13}
\end{figure}

\subsection{More Visualization Results}

To test the effectiveness of our proposed model on cross-database information, we used our model pre-trained on the GRID dataset to generate videos on the TCD-TIMIT test set. Fig.~\ref{fig.12} shows the results generated for different speakers using the same audio. The content of audio clip is the word "one.” We can see that the mouth movements produced by our model match the word well.

When no one was speaking, the mouth should not perform any movement and be consistent with the input image. We used audio with no one speaking to test our proposed model, which has only slight noise. Fig.~\ref{fig.13} shows the generation results for three different subjects using silent audio. Speakers were taken from the GRID test set. Although the mouth of the input face was open, the video generated by our method was also consistent with the input image, which demonstrates that our model is robust.

\section{Conclusion}

In this study, we have proposed a novel talking head generation system that can effectively integrate multimodal features and use both audio and speech-related facial AUs as driving information to drive talking head generation. The proposed dilated non-causal temporal convolutional self-attention network (TCSAN) was designed to promote the relationship learning of cross-modal representation and maintain the temporal relationship between consecutive frames. The proposed audio-to-AU module was used to obtain speech-related AU information. The AU classifier was used to ensure that the generated frames contained correct AU information. We conducted extensive experiments for quantitative and qualitative evaluation. We used an ablation study to verify each component's contribution to our model. The experimental results on the GRID dataset and TCD-TIMIT dataset demonstrated the superiority of the proposed approach over state-of-the-art methods in terms of both image quality and lip-sync accuracy. We think that the use of AU information can provide a reference for the task of talking head generation. Moreover, our TCSAN module also has strong applicability in other fields. We believe it can be extended to other cross-modal tasks to integrate multi-modal features, or applied to other sequence modeling tasks. 

Although our model can generate high-quality frontal talking heads, it cannot generate talking heads with specific emotion. In addition, the audio input to the model includes not only the information related to the speech content, but also other redundant information such as the identity of the speaker. In future work, we will focus on generating talking heads with specific emotion by introducing emotion-related information, and remove the redundant information irrelevant to the speech content in the audio. In these way, the authenticity of the synthesized talking head video can be further improved.

\section{Acknowledgement}
This work is supported by National Natural Science Foundation of China (No.61503277, No.61771333). We gratefully acknowledge the support of NVIDIA Corporation with the donation of the Titan V GPU used for this research.
\bibliography{reference}
\end{document}